\documentclass{acmtrans2f}

\usepackage{graphicx}
\usepackage{amssymb}

\def\qed{\relax\ifmmode\hskip2em \Box\else\unskip\nobreak\hskip1em
$\Box$\fi}

\newcommand{\myref}[1]{$(\ref{#1})$}

\newcommand{\rg}[2]{{[#1,#2]}}

\newcommand{\fuzzy}[3]{\langle{#1} {#2} {#3}\rangle}

\newcommand{\crisp}[1]{\mbox{$\sharp{(#1)}$}}

\newcommand{\mytrans}[1]{\mbox{$\overline{#1}$}}
\newcommand{\myStrans}[1]{\mbox{${\tt S}\crisp{#1}$}}

\newcommand{\langL}{\mbox{${\mathcal L}$}}
\newcommand{\langLf}{\mbox{${\mathcal L}^{f}$}}

\newcommand{\bottomf}{\perp}
\newcommand{\topf}{\top}

\newcommand{\andf}{\wedge}
\newcommand{\orf}{\vee}
\newcommand{\notf}{\neg}

\newcommand{\weakmod}{\mbox{$|\!\!\!\approx$}}

\newcommand{\entailtv}{\models_{2}}

\newcommand{\entailf}{\weakmod}
\newcommand{\notentailf}{\,\mbox{$\not \!\! \entailf$}}
\newcommand{\entailfv}{\models_{4}}

\newcommand{\equivf}{\cong}

\newcommand{\T}{{\tt T}}
\newcommand{\NT}{{\tt N\!T}}

\newcommand{\miff}{\mbox{iff}}

\newcommand{\ii}[1]{\mbox{$(#1)$}}

\newcommand{\bfl}{\begin{flushleft}}
\newcommand{\efl}{\end{flushleft}}

\newcommand{\ie}{i.e.}
\newcommand{\eg}{ e.g.}

\newcommand{\truef}[1]{ {t \in \highi{#1}} }
\newcommand{\falsef}[1]{ {f \in \highi{#1}} }
\newcommand{\nottruef}[1]{ {t \not \in \highi{#1}} }
\newcommand{\notfalsef}[1]{ {f \not \in \highi{#1}} }

\newtheorem{theorem}{Theorem}[section]

\newtheorem{corollary}[theorem]{Corollary}
\newtheorem{proposition}[theorem]{Proposition}

\newdef{definition}[theorem]{Definition}
\newdef{example}[theorem]{Example}
\newdef{remark}[theorem]{Remark}

\newcommand{\algocomment}[1]
{{\tt /* #1 */}}

\newcommand{\nd}{\noindent}

\newcommand{\ruleal}[3]
{ \raisebox{2ex}[0pt][0pt]{{\scriptsize #1 }   }
\begin{tabular}[b]{c}
   $#2$ \\ \hline
   $#3$
\end{tabular}
}

\newcommand{\rulebl}[4]
{ \raisebox{2ex}[0pt][0pt]{{\scriptsize #1 }   }   
\begin{tabular}[b]{c|c}
\multicolumn{2}{c}{$#2$} \\ \hline
   $#3$ & $#4$ 
\end{tabular}
}

\newcommand{\FKB}{{\Sigma}}

\newcommand{\I}{{\mathcal I}}

\newcommand{\highi}[1]{{#1}^{\mathcal I} }
\newcommand{\highv}[2]{{#1}^{#2} }

\usepackage{latexsym}

\newif\iftodo
 \todotrue      

\firstfoot{}
\runningfoot{}
\title{On the relationship between fuzzy logic and 
    four-valued relevance logic}
\author{UMBERTO STRACCIA \\
I.E.I. - C.N.R.}   

\begin{abstract}
In fuzzy propositional logic, to a proposition a partial truth in 
$[0,1]$ is assigned.  It is well known that under certain 
circumstances, fuzzy logic collapses to classical logic.  In this 
paper, we will show that under dual conditions, fuzzy logic collapses 
to four-valued (relevance) logic, where propositions have truth-value 
{\tt true,false, unknown}, or {\tt contradiction}.  As a consequence, 
fuzzy entailment may be considered as ``in between'' four-valued 
(relevance) entailment and classical entailment.
\end{abstract}



  
\begin{document}

\begin{bottomstuff}
Author's address: U. Straccia, I.E.I - C.N.R.,
Via Alfieri, 1 I-56010 San Giuliano (Pisa), ITALY. 
\newline
E-mail: straccia@iei.pi.cnr.it. 
\end{bottomstuff}

\maketitle


\section{Introduction}

\nd Since the introduction of fuzzy sets by \citeN{Zadeh65}, an 
impressive work has been carried out around them, not least the 
numerous studies on fuzzy logic.  In classical set theory, membership 
of a subset $S$ of the universe of objects $U$, is often viewed as a 
(crisp) characteristic function $\mu_{S}$ from $U$ to $\{0,1\}$ (called, 
\emph{valuation set}) such that

\[
\begin{array}{lcl}
\mu_{S}(u) & = & \left\{
                     \begin{array}{lcl}
		       1 & \miff & u \in S \\	 
		       0 & \miff & u \not\in S .	 
		     \end{array}
		 \right .
\end{array}
\]

\nd In fuzzy set theory, the valuation set is allowed to be the real 
interval $[0,1]$ and $\mu_{S}(u)$ is called the \emph{grade of 
membership\/}.  The closer the value to $1$, the more $u$ belongs to 
$S$.  Of course, $S$ is a subset of $U$ that has no sharp boundary.

When we switch to fuzzy propositional logic, the notion of grade of 
membership of an element $u$ in an universe $U$ with respect to a 
fuzzy subset $S$ over $U$ is regarded as the \emph{truth-value\/} of 
the proposition \emph{``u is $S$''}.  

In this paper we will consider a fuzzy propositional logic in which 
expressions are boolean combinations of simpler expressions of 
type $\fuzzy{A}{\geq}{n}$ and $\fuzzy{A}{\leq}{n}$, where $A$ is a 
propositional statement having a truth-value in $[0,1]$.  Both express 
a constraint on the truth-value of $A$, \ie \ a lower bound and an 
upper bound, respectively (see, \eg\ \cite{ChenJ96,Straccia00a}).  
While it is well-known that the fuzzy entailment relation, $\entailf$, 
is bounded \emph{upward} by classical entailment, $\entailtv$, \ie\ 
there cannot be fuzzy entailment without classical entailment, in this 
paper we will establish that the fuzzy entailment relation is bounded 
\emph{below} by the four-valued (relevance) entailment relation 
described in 
\cite{Anderson75,Belnap77a,Dunn86,Levesque84a,Straccia97a}, in which 
propositions have a truth-value {\tt true,false, unknown}, or {\tt 
contradiction}.  As a consequence, fuzzy entailment is ``in between'' 
the four-valued logical entailment relation $\entailfv$ and the 
classical two-valued logic entailment relation $\entailtv$.

We proceed as follows.  In the next section we introduce syntax, 
semantics of the fuzzy propositional logic considered, give main 
definitions, describe some basic properties and a decision procedure.  
In Section~\ref{fvsection} we will present the four-valued 
propositional logic considered in this paper, describe its properties 
and present a decision procedure.  Section~\ref{therelation} is the 
main part of this paper where the relations among fuzzy logic, 
four-valued logic and classical two-valued logic are described.  
Section~\ref{concl} concludes.


\section{A fuzzy propositional logic} \label{fuzzysection}

\subsection{Syntax and semantics}


\nd Our logical language has two parts.  At the \emph{objective 
level}, let \langL\ be the language of propositional logic, with 
connectives $\andf, \orf$, $\notf$, and the logical constants 
$\bottomf$ (false) and $\topf$ (true).  We will use metavariables $A, 
B, C, \ldots$ and $p,q,r,\ldots$ for propositions and propositional 
letters, respectively.\footnote{In the following, all metavariables 
could have an optional subscript and superscript.} $\bottomf, \topf$, 
letters and their negations are called {\em literals\/} (denoted $l$).  
As we will see below, propositions will have a truth-value in 
$\rg{0}{1}$.

At the \emph{meta level}, let \langLf\ be the language of {\em meta 
propositions\/} (denoted by $\psi)$.  \langLf\ consists of {\em meta 
atoms}, \ie \ expressions of type $\fuzzy{A}{\geq}{n}$ and 
$\fuzzy{A}{\leq}{n}$, where $A$ is a proposition in \langL\ and $n \in 
[0,1]$, the connectives $\andf, \orf$, $\notf$ and the logical 
constants $\bottomf$ and $\topf$.  Essentially, a meta-atom 
$\fuzzy{A}{\leq}{n}$ constrains the truth-value of $A$ to be less or 
equal to $n$ (similarly for $\geq$).  But, unlike \cite{Pavelka79} 
where the truth-value of $\fuzzy{A}{\leq}{n}$ can be any number in 
$[0,1]$, in our case $\fuzzy{A}{\geq}{n}$ and $\fuzzy{A}{\leq}{n}$ 
will have the truth-value $0$ or $1$.  Furthermore, a {\em meta 
letter\/} is a meta atom of the form $\fuzzy{p}{\geq}{n}$ and 
$\fuzzy{p}{\leq}{n}$, where $p$ is a propositional letter.  
$\bottomf,\topf$, meta letters and their negations are called {\em 
meta literals}.  A \emph{meta proposition} is then any $\andf, \orf$, 
$\notf$ combination of meta atoms.  For instance, $(\notf \fuzzy{r 
\andf s}{\leq}{0.6} \orf \fuzzy{p \orf q}{\geq}{0.2}) \andf \fuzzy{r 
\andf s}{\leq}{0.6}$ is a meta proposition, while 
$(\fuzzy{p}{\leq}{0.3} \geq 0.4)$ is not.  We will use 
$\fuzzy{A}{<}{n}$ as a short form of $\notf \fuzzy{A}{\geq}{n}$ and 
similarly for $\fuzzy{A}{>}{n}$; likewise, $\fuzzy{A}{=}{n}$ is a 
short form for $\fuzzy{A}{\leq}{n} \andf \fuzzy{A}{\geq}{n}$.  The 
meta letter $\fuzzy{p}{\geq}{n}$ is {\em non-trivial\/} if $n > 0$, 
and similarly for $\fuzzy{p}{\leq}{n}$.  The meta letter 
$\fuzzy{p}{\geq}{1}$ corresponds to the classical letter $p$ ($p$ is 
true), and $\fuzzy{p}{\leq}{0}$ corresponds to the classical literal 
$\notf p$ ($p$ is false).  Therefore, \langLf\ contains ${\mathcal 
L}$.

The classical definitions of \emph{Negation Normal Form\/} (NNF), 
\emph{Conjunctive Normal Form\/} (CNF) and \emph{Disjunctive Normal 
Form\/} (DNF) are easily extended to our context.  For instance, a 
meta proposition $\psi$ in \emph{negation normal form\/} is a 
combination of meta literals, using the connectives $\andf$ and 
$\orf$; a meta proposition $\psi$ in \emph{conjunctive normal form\/} 
is a conjunction of disjunction of meta literals.  Similarly for the 
DNF case.

From a semantics point of view, an interpretation $\I$ is a mapping 
$\highi{(\cdot)}$ from propositional letters into $\rg{0}{1}$. 
We extend $\I$ to propositions via the usual $\min,\max$ and 
1-complement functions:
$\highi{\topf} = 1$, 
$\highi{\bottomf} =  0$,
$\highi{(\notf A)} = 1- \highi{A}$,
$\highi{(A \andf B)} = \min\{\highi{A}, \highi{B} \}$,
$\highi{(A \orf B)} = \max\{\highi{A}, \highi{B} \}$.

Given an interpretation $\I$ we will assign a boolean truth-value 
in $\{0,1\}$ to each meta atom in the obvious way: namely,

\[
\begin{array}{lcl}
\highi{\fuzzy{A}{\geq}{n}}  =  1 & \mbox{ iff } & \highi{A} \geq n, \mbox{ and } \\
\highi{\fuzzy{A}{\leq}{n}}  =  1 & \mbox{ iff } & \highi{A} \leq n .
\end{array}
\]

\nd
Finally, we assign a boolean truth-value to each meta proposition like 
$\fuzzy{A}{\geq}{n_{1}} \orf \fuzzy{B}{\leq}{n_{2}}$ using the 
classical method of combining truth-values and we say that an 
interpretation $\I$ \emph{satisfies\/} a meta proposition $\psi$ iff 
$\highi{\psi} = 1$; in that case, we will say that $\I$ is a 
\emph{model\/} of $\psi$.

A {\em meta theory\/} (denoted by $\Sigma$) is a finite set of meta 
propositions.  Given an interpretation $\I$ and a meta theory 
$\Sigma$, we say that $\I$ \emph{satisfies\/} $\Sigma$ if $\I$ 
satisfies each $\psi \in \Sigma$; in that case we say that $\I$ is a 
\emph{model\/} of $\Sigma$.  We say that a meta theory $\Sigma$ 
\emph{entails\/} a meta proposition $\psi$ if every model of $\Sigma$ 
is a model of $\psi$; this is denoted by $\Sigma \entailf \psi$.  A 
meta proposition $\psi$ is \emph{valid\/} if it is entailed by the 
empty meta theory, \ie \ $\emptyset \entailf \psi$.  An example of 
valid meta proposition is $\fuzzy{p \orf \notf p}{\geq}{0.5}$.  Two 
propositions $A$ and $B$ are said to be \emph{equivalent\/} (denoted 
by $A \equiv B)$ if $\highi{A} = \highi{B}$, for each interpretation 
$\I$.  For example, $\notf (A \andf \notf B)$ is equivalent to $\notf 
A \orf B$.  The equivalence of two meta propositions, $\psi \equivf 
\psi'$ , is defined similarly.

Given a meta theory $\Sigma$ and a proposition $A$, it is of interest 
to compute $A$'s best lower and upper truth-value bounds 
\cite{Straccia00a}.  To this end we define the {\em least upper 
bound\/} and the {\em greatest lower bound\/} of $A$ with respect to 
$\Sigma$ (written $lub(\Sigma, A)$ and $glb(\Sigma, A)$, respectively) 
as $lub(\Sigma, A) = \inf\{n : \Sigma \entailf \fuzzy{A}{\leq}{n} \}$ 
and $glb(\Sigma, A) = \sup\{n : \Sigma \entailf \fuzzy{A}{\geq}{n} 
\}$.

\subsection{Some basic properties} \label{basicp}

\nd In order to make our paper self-contained, we recall some 
properties of the logic \langLf, which will be of use (see 
\cite{Straccia00a}).  The first ones are straightforward: for any 
proposition $A,B$ and $C$, $\notf \topf \equiv \bottomf$, $ A \andf 
\topf \equiv A$, $A \orf \topf \equiv \topf$, $A \andf \bottomf \equiv 
\bottomf$, $ A \orf \bottomf \equiv A$, $\notf \notf A \equiv A$, 
$\notf (A \andf B) \equiv \notf A \orf \notf B$, $\notf (A \orf B) 
\equiv \notf A \andf \notf B$, $(A \andf (B \orf C)) \equiv (A \andf 
B) \orf (A \andf C)$ and $(A \orf (B \andf C)) \equiv (A \orf B) \andf 
(A \orf C)$.  It can be verified that each proposition may easily be 
transformed, by preserving equivalence, into either $\topf$, 
$\bottomf$ or a proposition in NNF, CNF and DNF in which neither 
$\topf$ nor $\bottomf$ occur.  Please note, we do not have $A \andf 
\notf A \equiv \bottomf$.  In general we can only say that $\highi{(A 
\andf \notf A)} \leq 0.5$, for any interpretation ${\mathcal I}$ and 
similarly $\highi{(A \orf \notf A)} \geq 0.5$.

Concerning meta propositions, as meta propositions have a boolean 
truth-value, we have the equivalencies of classical propositional 
logic, \eg \ $\psi \andf \topf \equivf \psi$, $\notf \psi \andf \psi 
\equivf \bottomf$, as well as

\begin{eqnarray}
    \fuzzy{\topf}{\geq}{n} & \equivf & \topf
                      \label{eq:equivtopfgeq}  \nonumber  \\
    \fuzzy{\topf}{\leq}{n} & \equivf & \left \{ \begin{array}{lcl}
                                                 \topf & \mbox{if $n=1$} \\
                                                 \bottomf & \mbox{otherwise}
					       \end{array}	 
                                      \right.  \label{eq:equivtopfleq} \nonumber \\
     \fuzzy{p}{\geq}{0} & \equivf & \topf
                      \label{eq:equivpgeq} \nonumber \\ 
     \fuzzy{p}{\leq}{1} & \equivf & \topf
                      \label{eq:equivpleq}  \nonumber \\ 
    \fuzzy{\notf A}{\geq}{n} & \equivf & \fuzzy{A}{\leq}{1 - n}
                      \label{eq:equivnotf}  \\
    \fuzzy{A \andf B}{\geq}{n} & \equivf & \fuzzy{A}{\geq}{n} \andf \fuzzy{B}{\geq}{n}
                      \label{eq:equivandf} \nonumber \\
    \fuzzy{A \orf B}{\geq}{n} & \equivf & \fuzzy{A}{\geq}{n} \orf \fuzzy{B}{\geq}{n}
                      \label{eq:equivorf} \nonumber
\end{eqnarray}

\nd and likewise for the cases $\leq, <$ and $>$.  Therefore, each 
meta proposition may easily be transformed by, preserving equivalence, 
into $\topf$, $\bottomf$ or into a meta proposition in NNF, CNF and 
DNF in which neither $\topf$ nor $\bottomf$ occur.  Since $\Sigma 
\entailf \topf, \Sigma \notentailf \bottomf$ (unless $\Sigma$ is 
unsatisfiable), $glb(\Sigma,\topf) = 1, glb(\Sigma,\bottomf) = 0, 
lub(\Sigma,\topf) = 1, lub(\Sigma,\bottomf) = 0$, $\Sigma \cup 
\{\topf\}$ and $\Sigma$ share the same set of models and 
$\Sigma\cup\{\bottomf\}$ is unsatisfiable, for the rest of the paper, 
if not stated otherwise, \emph{we will always assume that meta 
propositions are always in NNF in which neither trivial meta letters 
nor $\topf$ nor $\bottomf$ occur\/}.

As showed in \cite{Straccia00a}, there is a strict relation between 
meta propositions and classical propositions.  Let us consider the 
following transformation $\crisp{\cdot}$ of meta propositions into 
propositions, where $\crisp{\cdot}$ takes the ``crisp'' 
propositional part of a meta proposition:

\begin{eqnarray*}
    \crisp{\fuzzy{p}{\geq}{n}} & \mapsto  & p
    \label{eq:crispp}  \\
    \crisp{\fuzzy{p}{\leq}{n}} & \mapsto  & \notf p
    \label{eq:crispnotp}  \\
    \crisp{\notf \psi} & \mapsto  & \notf \crisp{\psi}
    \label{eq:crispnot}   \\
    \crisp{\psi_{1} \andf \psi_{2}} & \mapsto  & 
              \crisp{\psi_{1}} \andf \crisp{\psi_{2}}
    \label{eq:crispand}  \\
    \crisp{\psi_{1} \orf \psi_{2}} & \mapsto  & 
              \crisp{\psi_{1}} \orf \crisp{\psi_{2}} .
    \label{eq:crispor}
\end{eqnarray*}

\nd Further, for a meta theory $\Sigma$, $\crisp{\Sigma}= 
\{\crisp{\psi}: \psi \in \Sigma\}$.  Then the following proposition 
holds.

\begin{proposition}[\cite{Straccia00a}] \label{prop1}
Let $\Sigma$ be a meta theory and let $\psi$ be a meta proposition: 
\begin{enumerate}
     \item if $\Sigma$ is unsatisfiable then $\crisp{\Sigma}$ is 
     classically unsatisfiable;

    \item if $\Sigma \entailf \psi$ then $\crisp{\Sigma} \entailtv 
    \crisp{\psi}$, where $\entailtv$ is classical entailment.   
\end{enumerate}
\end{proposition}

\nd Proposition~\ref{prop1} states that {\em there cannot be 
entailment without classical entailment}. In this sense $\entailf$ is 
correct with respect to $\entailtv$.

\begin{example} \label{ex1}
Let $\Sigma$ be the set 
$\Sigma = \{
\fuzzy{p}{\geq}{0.8} \orf  \fuzzy{q}{\leq}{0.3}, 
\fuzzy{p}{\leq}{0.3} \}$. Let $\psi$ be $\fuzzy{q}{\leq}{0.6}$. 
It follows that 
$\crisp{\Sigma}= \{p \orf  \notf q, \notf p \}$.
It is easily verified that $\Sigma \entailf \fuzzy{q}{\leq}{0.6}$ and 
that $\crisp{\Sigma} \entailtv \notf q$, thereby 
confirming Proposition~\ref{prop1}. 
\end{example}

\nd The converse of Proposition~\ref{prop1} does not hold in the 
general case.

\begin{example} \label{ex2}
Let $\Sigma$ be the set 
$\Sigma = \{ \fuzzy{p}{\leq}{0.5} \orf \fuzzy{q}{\geq}{0.6}, 
\fuzzy{p}{\geq}{0.3} \}$.  
It follows that 
$\crisp{\Sigma} = \{\notf p \orf q, p\}$.  
It is easily verified that $\crisp{\Sigma} \entailtv q$, but 
$\Sigma \notentailf \fuzzy{q}{\geq}{n}$, for all $n >0$.

\end{example}

\nd The following result establishes the converse of 
Proposition~\ref{prop1}.  It directly relates to a similar result 
described in \cite{Lee72}.  We say that a meta proposition $\psi$ is 
\emph{normalised\/} iff for each meta literal $\psi'$ occurring in 
$\psi$,

\begin{enumerate}
    \item  if $\psi'$ is $\fuzzy{p}{\geq}{n}$ then $n > 0.5$; 

    \item  if $\psi'$ is $\fuzzy{p}{\leq}{n}$ then $n < 0.5$;

    \item  if $\psi'$ is $\fuzzy{p}{>}{n}$ then $n \geq 0.5$;

    \item  if $\psi'$ is $\fuzzy{p}{<}{n}$ then $n \leq 0.5$.
\end{enumerate}

\begin{proposition}[\cite{Straccia00a}]\label{prop2}
Let $\Sigma$ be a meta theory and let $\psi$ be a meta proposition.  
Furthermore, we assume that each $\psi' \in \Sigma$ is normalised as 
well as is an equivalent NNF of $\notf \psi$.  Then

\begin{enumerate}
\item $\Sigma$ is satisfiable iff $\crisp{\Sigma}$ is classically satisfiable; 
\item $\Sigma \entailf \psi$ iff $\crisp{\Sigma} \entailtv 
\crisp{\psi}$, where $\entailtv$ is classical entailment.  
\end{enumerate}
\end{proposition}

\begin{example} \label{ex3}
Consider Example~\ref{ex1}.  An equivalent NNF of $\notf \psi$ is 
$\psi' = \fuzzy{q}{>}{0.6}$.  It is easily verified that both $\Sigma$ 
and $\psi$ are normalised.  Indeed, both $\Sigma \entailf \psi$ and 
$\crisp{\Sigma} \entailtv \crisp{\psi}$ hold.  On the other hand, 
in Example~\ref{ex2}, $\Sigma$ is not normalised, \eg \ for 
$\fuzzy{p}{\geq}{0.3}$ we have $0.3 < 0.5$.

\end{example}

\nd Dually to normalisation, we say that 
a meta proposition $\psi$ is \emph{sub-normalised\/} iff for each meta 
literal $\psi'$ occurring in $\psi$,

\begin{enumerate}
    \item  if $\psi'$ is $\fuzzy{p}{\geq}{n}$ then $n \leq 0.5$; 

    \item  if $\psi'$ is $\fuzzy{p}{\leq}{n}$ then $n \geq 0.5$;

    \item  if $\psi'$ is $\fuzzy{p}{>}{n}$ then $n < 0.5$;

    \item  if $\psi'$ is $\fuzzy{p}{<}{n}$ then $n > 0.5$.
\end{enumerate}

\nd
Furthermore, for any letter $p$ and meta proposition $\psi$, let ($\max\emptyset = 
0, \min\emptyset = 1$):

\begin{eqnarray}
p^{\geq}_{\psi} & = & \max\{n: \fuzzy{p}{\geq}{n} \mbox{ occurs in } \psi 
\} \label{sn1} \\
p^{>}_{\psi}& = & \max\{n: \fuzzy{p}{>}{n} \mbox{ occurs in } \psi \}    \label{sn2} \\
p^{\leq}_{\psi} & = & \min\{n: \fuzzy{p}{\leq}{n} \mbox{ occurs in } \psi\}  \label{sn3} \\
p^{<}_{\psi} & = & \min\{n: \fuzzy{p}{<}{n} \mbox{ occurs in } \psi \} .  
\label{sn4}
\end{eqnarray}

\nd For any $p$ and $\psi$, $p^{\geq}_{\psi}, p^{>}_{\psi}$ and $p^{\leq}_{\psi}, 
p^{<}_{\psi}$, determine the greatest lower bound and the least upper 
bound which $p$'s truth value has to satisfy, respectively. We extend 
the above definition to the case of meta theories as follows:

\begin{eqnarray}
p^{\geq}_{\Sigma} & = & \max\{p^{\geq}_{\psi}: \psi \in \Sigma \} \label{ns1} \\
p^{>}_{\Sigma}& = & \max\{p^{>}_{\psi}: \psi \in \Sigma \}    \label{ns2} \\
p^{\leq}_{\Sigma} & = & \min\{p^{\leq}_{\psi}: \psi \in \Sigma\}  \label{ns3} \\
p^{<}_{\Sigma} & = & \min\{p^{<}_{\psi}: \psi \in \Sigma\} .  \label{ns4}
\end{eqnarray}

The following proposition holds.

\begin{proposition} \label{propa1}
Let $\psi$ be a sub-normalised meta proposition in NNF.  Then $\psi$ is 
satisfiable.  
\end{proposition}

\begin{proof}
For any letter $p$ consider $p^{\geq}_{\psi}, p^{>}_{\psi}, 
p^{\leq}_{\psi}$ and $p^{<}_{\psi}$.  Since $\psi$ is 
sub-normalised  it follows that for each letter $p$, there is $\epsilon_{p} 
\geq 0$ such that

\begin{equation} \label{srestr}
\underline{p} = \max\{p^{\geq}_{\psi},p^{>}_{\psi} + \epsilon_{p}\} \leq 
\min\{p^{\leq}_{\psi},p^{<}_{\psi} - \epsilon_{p}\} =  \overline{p}
\end{equation}

\nd \ie, for each $p$, its greatest lower bound constraint is less or 
equal than its least upper bound constraint.  Now, let $\I$ be an 
interpretation such that

\begin{enumerate}
\item $\highi{\topf} = 1$ and $\highi{\bottomf} = 0$;
\item $\highi{p} = \underline{p}$, for all letters $p$ .
\end{enumerate}

\nd We will show on induction on the number of connectives of $\psi$ 
that $\I$ is an interpretation satisfying $\psi$.
\begin{description}
\item[$\psi$ is a meta letter] \mbox{ \ } 
\begin{enumerate}
    \item Suppose $\psi$ is a meta letter $\fuzzy{p}{\geq}{n}$.  
    By definition, $n \leq \underline{p}$ and, thus, $\I$ satisfies 
    $\fuzzy{p}{\geq}{n}$. 
    
    \item Suppose $\psi$ is a meta letter $\fuzzy{p}{\leq}{n}$.  By 
    definition, $n \geq \overline{p} \geq \underline{p}$ and, thus, 
    $\I$ satisfies $\fuzzy{p}{\leq}{n}$.
\end{enumerate}

\item[Induction step] \mbox{ \ } 
\begin{enumerate}
    \item Suppose $\psi$ is a meta proposition $\psi_{1} \andf 
    \psi_{2}$.  By induction on 
    $\psi_{1}$ and $\psi_{2}$, $\I$ satisfies 
    $\psi_{1}$ and $\psi_{2}$ and, thus, $\I$ satisfies $\psi$.
    
    \item The cases $\orf$ is similar. 
\end{enumerate}
\end{description}
\end{proof}

\nd The above property is easily generalised to sub-normalised meta 
theories.  We say that a meta theory $\Sigma$ is \emph{sub-normalised} 
iff each element of it is.

\begin{corollary} \label{cora1}
Let $\Sigma$ be a sub-normalised meta theory in NNF.  Then $\Sigma$ is 
satisfiable.  
\end{corollary}

\begin{proof}
Similarly to Proposition~\ref{propa1}, for any letter $p$, 
consider $p^{\geq}_{\Sigma}, p^{>}_{\Sigma}, p^{\leq}_{\Sigma}$ and 
$p^{<}_{\Sigma}$.  Since $\Sigma$ is sub-normalised, it follows that 
for each letter $p$, there is $\epsilon_{p} \geq 0$ such that
$
\underline{p} = \max\{p^{\geq},p^{>} + \epsilon_{p}\} \leq 
\min\{p^{\leq},p^{<} - \epsilon_{p}\} =  \overline{p}
$.
Now, let $\I$ be an interpretation such that

\begin{enumerate}
\item $\highi{\topf} = 1$ and $\highi{\bottomf} = 0$;
\item $\highi{p} = \underline{p}$, for all letters $p$ .
\end{enumerate}

\nd
It is easily verified that $\I$ satisfies $\Sigma$. 
\end{proof}

\nd
As it happens for classical entailment, entailment in \langLf\ 
can be reduced to satisfiability checking: indeed, for a meta 
theory $\Sigma$ and a meta proposition $\psi$

\begin{equation}
    \begin{array}{lcl}
       \Sigma \entailf \psi & \miff & \Sigma \cup \{ \notf \psi\} 
       \mbox{ is unsatisfiable.}
    \end{array}
    \label{eq:unsat}
\end{equation}

\nd
We conclude this section by showing that the computation of the least 
upper bound can be reduced to the computation of the greatest lower 
bound.  Let $\Sigma$ be a meta theory and let $A$ be a proposition.  By 
\myref{eq:equivnotf}, $\fuzzy{A}{\leq}{n} \equiv \fuzzy{\notf 
A}{\geq}{1-n}$ holds and, thus,
$\Sigma \entailf \fuzzy{A}{\leq}{n}$ iff
$\Sigma \entailf \fuzzy{\notf A}{\geq}{1-n}$ holds.
Therefore, 

\[
\begin{array}{lcl}
1-lub(\Sigma,A) 
& = & 1 - \inf\{n: \Sigma \entailf \fuzzy{A}{\leq}{n} \} \\
& = & \sup\{1-n: \Sigma \entailf \fuzzy{A}{\leq}{n} \} \\
& = & \sup\{n: \Sigma \entailf \fuzzy{A}{\leq}{1-n} \} \\
& = & \sup\{n: \Sigma \entailf \fuzzy{\notf A}{\geq}{n} \} \\
& = & glb(\Sigma,\notf A) 
\end{array}
\]

\nd
and, thus,

\begin{equation}\label{lubglb}
\begin{array}{lcl}
lub(\Sigma,A) & = & 1 - glb(\Sigma,\notf A) ,
\end{array}
\end{equation}

\nd
\ie \ the $lub$ can be determined through the $glb$ (and vice-versa).

\nd In \cite{Straccia00a} a simple method has been developed in order 
to compute the $glb$.  The method is based on the fact that from 
$\Sigma$ it is possible to determine a finite set $N^{\Sigma} \subset 
[0,1]$, where $|N^{\Sigma}|$ is $O(|\Sigma|)$, such that 
$glb(\Sigma,A) \in N^{\Sigma}$.  Therefore, $glb(\Sigma,A)$ can be 
determined by computing the greatest value $n \in N^{\Sigma}$ such 
that $\Sigma \entailf \fuzzy{A}{\geq}{n}\}$.  An easy way to search 
for this $n$ is to order the elements of $N^{\Sigma}$ and then to 
perform a binary search among these values.

\begin{proposition}[\cite{Straccia00a}] \label{prop44}
Let $\Sigma$ be a meta theory.  Then $glb(\Sigma,A) \in N^{\Sigma}$, 
where

\begin{equation} \label{eq:nsigma}
\begin{array}{lcl}
     N^{\Sigma} & = & \{ 0,0.5,1 \} \ \cup \\ 
                &   & \{n:  \fuzzy{p}{\geq}{n} \mbox{ or } 
                            \fuzzy{p}{>}{n} \mbox{ occurs in } \Sigma \} \ \cup \\
                &   & \{1-n: \fuzzy{p}{\leq}{n} \mbox{ or } 
                             \fuzzy{p}{<}{n} \mbox{ occurs in } \Sigma\} . 
\end{array}
\end{equation}
\mbox{ \ }  
\end{proposition}
    
\nd For instance, for the meta theory in Example~\ref{ex1}, 
$N^{\Sigma}$ is given by $\{0, 0.5, 1\} \cup \{0.8, 0.7\}$.  The value 
of $glb(\Sigma,A)$ can, thus, be determined in $O(\log |N^{\Sigma}|)$ 
entailment tests.

Note that, since for a proposition $A$, $\fuzzy{A}{<}{0.5}$ is 
normalised, it follows from Proposition~\ref{prop2} that $\entailtv A$ iff 
$\entailf \fuzzy{A}{\geq}{0.5}$, \ie\ the truth-value of a classical 
propositional tautology is greater or equal than $0.5$.  But, by 
Proposition~\ref{prop44}, $glb(\emptyset,A) \in \{0,0.5\}$ and, thus, 
$\entailtv A$ iff $glb(\emptyset,A) = 0.5$, \ie \ a classical tautology 
has $0.5$ as its greatest truth-value lover bound.

\subsection{Decision procedure in \langLf} \label{autoreas}

\nd In this section we will present a procedure for deciding the main 
problem within \langLf: deciding whether a meta theory $\Sigma$ is 
satisfiable or not (by \myref{eq:unsat}, the entailment problem is 
solved too).  We call it the \emph{fuzzy SAT problem\/} in order to 
distinguish it from the classical SAT problem.

We recall here a simplified version of the decision procedure proposed 
in \cite{Straccia00a}.  Given two meta propositions $\psi_{1}$ and 
$\psi_{2}$ we say that \ii{i} $\psi_{1}$ \emph{subsumes\/} $\psi_{2}$ 
(denoted by $subs(\psi_{1}, \psi_{2})$) iff $\psi_{1} \entailf 
\psi_{2}$; and that \ii{ii} $\psi_{1}$ and $\psi_{2}$ are 
\emph{pairwise contradictory\/} (denoted by $ctd(\psi_{1}, \psi_{2})$) 
iff $\psi_{1} \entailf \notf \psi_{2}$.  For instance, 
$\fuzzy{p}{\geq}{0.3} \orf \fuzzy{q}{\leq}{0.6}$ subsumes 
$\fuzzy{p}{\geq}{0.2} \orf \fuzzy{q}{\leq}{0.9}$, while 
$\fuzzy{p}{\geq}{0.3} \orf \fuzzy{q}{\leq}{0.6}$ and 
$\fuzzy{p}{\leq}{0.2} \andf \fuzzy{q}{\geq}{0.7}$ are pairwise 
contradictory.  Since $\psi_{1} \entailf \notf \psi_{2}$ iff $\psi_{2} 
\entailf \notf \psi_{1}$ it follows that $ctd(\cdot, \cdot)$ is 
symmetric.  By definition, $ctd(\psi_{1}, \psi_{2})$ \miff\ 
$subs(\psi_{1},\notf \psi_{2})$ holds which relates $ctd(\cdot, 
\cdot)$ to $subs(\cdot, \cdot)$.  If $\psi_{1}$ and $\psi_{2}$ are two 
meta literals, it is quite easy to check whether $subs(\psi_{1}, 
\psi_{2})$ holds, as shown in Table~\ref{subs}, on the left.  Each 
entry in the table specifies the condition under which $\psi_{1}$ 
subsumes $\psi_{2}$.
\begin{table}
    \caption{On the left: $\psi_{1}$ subsumes $\psi_{2}$.  On the right: 
$\psi_{1}$ and $\psi_{2}$ pairwise contradictory.} \label{subs}
\begin{center}
\begin{tabular}{c}
\begin{tabular}{cc} 
\begin{tabular}{|c||c|c|c|c|} \hline 
    $\psi_{1}$ & \multicolumn{4}{c|}{$\psi_{2}$}     \\ \hline 
          & $\fuzzy{p}{\geq}{m}$ & $\fuzzy{p}{>}{m}$
	  & $\fuzzy{p}{\leq}{m}$ & $\fuzzy{p}{<}{m}$  \\ \hline \hline
    $\fuzzy{p}{\geq}{n}$  & $n \geq m$ & $n > m$ & $\times$ & $\times$  \\ \hline
    $\fuzzy{p}{>}{n}$     & $n \geq m$ & $n \geq m$ &$\times$ & $\times$ \\ \hline
    $\fuzzy{p}{\leq}{n}$  & $\times$ & $\times$ &$n \leq m$ & $n < m$ \\ \hline
    $\fuzzy{p}{<}{n}$     & $\times$ & $\times$ & $n \leq m$ & $n \leq m$ \\ \hline
\end{tabular}
&
\begin{tabular}{|c||c|c|} \hline
    $\psi_{1}$ & \multicolumn{2}{c|}{$\psi_{2}$}     \\ \hline
          & $\fuzzy{p}{\geq}{m}$ & $\fuzzy{p}{>}{m}$ \\ \hline \hline
$\fuzzy{p}{\leq}{n}$  & $n < m$ & $n \leq m$ \\ \hline
$\fuzzy{p}{<}{n}$     & $n \leq m$ & $n \leq m$ \\ \hline
\end{tabular}
\end{tabular}
\end{tabular}
\end{center}
\end{table}
We are now ready to specify the calculus.  The calculus is based 
on the following set of rules, ${\mathcal R}^{T} = \{(\bottomf), 
(\andf), (\orf)\}$, described in Table~\ref{calculus}.
\begin{table}
    \caption{Simple Tableaux inference rules for \langLf.}
\label{calculus}
\begin{center}
\begin{tabular}{c}
\begin{tabular}{cl}
$\ruleal{$(\bottomf)$}
{\psi, \psi'} 
{\bottomf}$ & 
\parbox{10cm}{where $\psi,\psi'$ are meta literals and 
$ctd(\psi,\psi')$} \\
\\
$\ruleal{$(\andf)$}
{\psi_{1} \andf \psi_{2}} 
{\psi_1, \psi_2}$ & \\
\\
$\rulebl{$(\orf)$}
{\psi_{1} \orf \psi_{2}} 
{\psi_1}
{\psi_2}$ &  \\
\end{tabular}
\end{tabular}
\end{center}
\end{table}
As usual, a deduction is represented as a tree, called {\em deduction 
tree}.  A branch $\phi$ in a deduction tree is closed iff it contains 
$\bottomf$.  A deduction tree is \emph{closed\/} iff each branch in it 
is closed.  With $\phi^{M}$ we indicate the set of meta propositions 
occurring in $\phi$.  A meta theory $\Sigma$ has a {\em refutation\/} 
iff each deduction tree is closed.  A branch ${\phi}$ is 
\emph{completed\/} iff it is not closed and no rule can be further 
applied to it.  A branch ${\phi}$ is \emph{open\/} iff it is not 
closed and not completed. 

Given a meta theory $\Sigma$, the procedure ${\sf SAT}(\Sigma)$ 
described in Figure~\ref{tabsat} determines whether $\Sigma$ is 
satisfiable or not.  ${\sf SAT}(\Sigma)$ starts from the root labelled 
$\Sigma$ and applies the rules until the resulting tree is either 
closed or there is a completed branch.  If the tree is closed, ${\sf 
SAT}(\Sigma)$ returns false, otherwise true and from the completed 
branch a model of $\Sigma$ can be build.  The set of not closed 
branches $\phi$ which may be expanded during the deduction is hold by 
$\Phi$.
\begin{figure}
{\bf procedure ${\sf SAT}(\Sigma)$}

\nd Convert each $\psi \in \Sigma$ into an equivalent NNF.
${\sf SAT}(\Sigma)$ starts from the root labelled $\Sigma$.  So, we 
initialise $\Phi$ with $\Phi = \{\phi\}$, where $\phi^{M} = \Sigma$.  
$\Phi$ is managed as a multiset, \ie \ there could be elements in 
$\Phi$ which are replicated.

\begin{enumerate}   
\item if $\Phi = \emptyset$ then return {\tt false} and exit;

\algocomment{all branches are closed and, thus, $\Sigma$ is unsatisfiable}

\item otherwise, select a branch $\phi \in \Phi$ and remove it from $\Phi$, \ie \ 
$\Phi \leftarrow \Phi \setminus \{\phi\}$;

\item try to apply a rule to $\phi$ with the following priority 
among the rules: $(\bottomf) \succ (\andf) \succ (\orf)$:

\begin{enumerate}
    \item if the $(\bottomf)$ rule is applicable to $\phi$ then go to 
    step 1.

    \item if the $(\andf)$ rule is applicable to $\phi$ then expand 
    $\phi$ by the application of the $(\andf)$ rule.  Let $\phi'$ be 
    the resulting branch.  If $\phi'$ is not closed then add it to 
    $\Phi$, \ie \ $\Phi \leftarrow \Phi \cup \{\phi'\}$; Go to step 1.

    \item if the $(\orf)$ rule is applicable to $\phi$ then expand 
    $\phi$ by the application of the $(\orf)$ rule.  Let $\phi_{1}$ and 
    $\phi_{2}$ be the resulting branches.  For each $\phi_{i}, i=1,2$, 
    if $\phi_{i}$ is not closed then add it to $\Phi$, \ie \ 
    $\Phi \leftarrow \Phi \cup \{\phi_{i}\}$. Go to step 1.

    \item otherwise, if no rule is applicable to $\phi$, then return 
    {\tt true} and exit.  
    
    \algocomment{$\phi$ is completed and, thus, $\Sigma$ is satisfiable}
    
\end{enumerate}
\end{enumerate}

{\bf end ${\sf SAT}$}

\caption{The procedure ${\sf SAT}$.} \label{tabsat}
\end{figure}

\begin{example} \label{ex4}
Let $\Sigma$ be the set 

\[
\begin{array}{lcl}
\Sigma & = & \{  
   \fuzzy{p}{\geq}{0.5} \orf (\fuzzy{q}{\geq}{0.4} \andf \fuzzy{u}{\geq}{0.6}), 
 \fuzzy{p}{\leq}{0.3} \}
\end{array}
\]

\nd
Figure~\ref{fig1} shows a deduction tree produced by ${\sf SAT}(\Sigma)$.
\begin{figure}
\begin{center}
\begin{tabular}{c}
\setlength{\unitlength}{0.5cm}
\begin{picture}(14,9) 
   \put(5,8){\makebox(6,1){$\{\fuzzy{p}{\geq}{0.5} \orf (\fuzzy{q}{\geq}{0.4} \andf \fuzzy{u}{\geq}{0.6}), 
 \fuzzy{p}{\leq}{0.3}  \}$}}
   \put(8,8){\line(-2,-1){4}}
   \put(8,8){\line(2,-1){4}}
   \put(3,5){\makebox(2,1){$\fuzzy{p}{\geq}{0.5}$}}
   \put(4,5){\line(0,-1){1}}
   \put(3,3){\makebox(2,1){$\bottomf$}}	    
   \put(3,1){\makebox(2,1){$\times$}}
   \put(11,5){\makebox(2,1){$\fuzzy{q}{\geq}{0.4} \andf \fuzzy{u}{\geq}{0.6}$}}
   \put(12,5){\line(0,-1){2}}
   \put(11,2){\makebox(2,1){$\fuzzy{q}{\geq}{0.4}$,
                            $\fuzzy{u}{\geq}{0.6}$,}}
   \put(11,0){\makebox(2,1){$\phi$}}
\end{picture}
\end{tabular}
\end{center}
\caption{Deduction tree for $\Sigma$.} \label{fig1}
\end{figure}
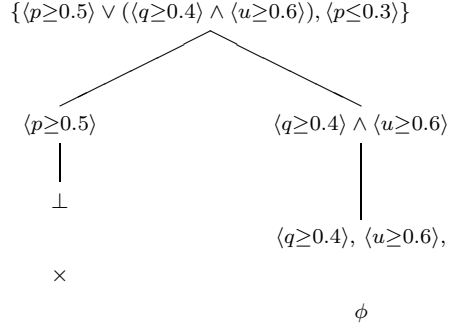
The branch on the left is closed, while the branch $\phi$ on 
the right is completed. Consider $\phi'^{M} \subseteq \phi^{M}$ where
$\phi'^{M}$ contains all the meta literals occurring in $\phi^{M}$, \ie \

\begin{eqnarray*}
\phi'^{M} & = & \{
   \fuzzy{p}{\leq}{0.3},\fuzzy{q}{\geq}{0.4}, \fuzzy{u}{\geq}{0.6}\} .
\end{eqnarray*}

\nd
From $\phi'^{M}$ a model $\I$ of $\Sigma$ can easily be build as follows:
$\highi{p}=0.3,\highi{q}=0.4$ and $\highi{u}=0.6$.

\end{example}

\nd The following 
proposition establishing correctness and completeness of the 
${\sf SAT}$ procedure.

\begin{proposition}[\cite{Straccia00a}] \label{satcomplete}
Let $\Sigma$ be a meta theory.  Then ${\sf SAT}(\Sigma)$ iff 
$\Sigma$ is satisfiable.  
\end{proposition}


\section{Four-valued propositional logic} \label{fvsection}

\nd The four-valued propositional logic we will rely on can be found in
\cite{Anderson75,Belnap77a,Dunn86,Levesque84a,Straccia97a,Straccia99a}. 
In the following we will describe briefly syntax, semantics, basic 
properties and a decision procedure for the four-valued entailment 
problem.

Expressions in four-valued propositional logic are propositions in 
which no $\topf$ and $\bottomf$ appear.  A theory is a set of propositions.  
From a semantics point of view, a {\em four-valued interpretation} 
$\I$ maps a proposition into an element of $2^{\{t,f\}} =\{\emptyset, 
\{t\}, \{f\}, \{t,f\}\}$.  The four truth-values, $\emptyset, \{t\}, 
\{f\}, \{t,f\}$ stand for {\tt unknown}, {\tt true}, {\tt false} and 
{\tt contradiction}, respectively.  Furthermore, $\I$ has to satisfy 
the following equations:

\begin{center}
\begin{tabular}{lcl}
 $t \in$ $\highi{(A \andf B)}$  & iff & $t \in$ $\highi{A}$ and $t \in$ $\highi{B}$; \\
 $f \in$ $\highi{(A \andf B)}$ & iff & $f \in$ $\highi{A}$ or $f \in$ 
 $\highi{B}$; \\
 $t \in$ $\highi{(A \orf B)}$ & iff & $t \in$ $\highi{A}$ or $t \in$ 
 $\highi{B}$; \\
 $f \in$ $\highi{(A \orf B)}$ & iff & $f \in$ $\highi{A}$ and $f \in$ 
 $\highi{B}$; \\
 $t \in$ $\highi{(\notf A)}$ & iff & $f \in$ $\highi{A}$ \\
 $f \in$ $\highi{(\notf A)}$ & iff & $t \in$ $\highi{A}$. 
\end{tabular}
\end{center}

\nd It is worth noting that a two-valued interpretation is just a 
four-valued interpretation $\I$ such that $\highi{p}$ $\in$ 
$\{\{t\},\{f\}\}$, for each letter $p$.  We might characterise the 
distinction between two-valued and four-valued semantics as the 
distinction between {\em implicit} and {\em explicit} falsehood: in a 
two-valued logic a formula is (implicitly) false in an interpretation 
iff it is not true, while in a four-valued logic this need not be the 
case.  Our truth conditions are always given in terms of belongings 
$\in$ (and never in terms of non belongings $\not \in$) of 
truth-values to interpretations.  Let $\I$ be a four-valued 
interpretation, let $A,B$ be two propositions and let $\Sigma$ be a 
theory: $\I$ {\em satisfies} ({\em is a model of}) $A$ iff $\truef{A}$; 
$\I$ \emph{satisfies} ({\em is a model of}) $\Sigma$ iff $\I$ is a 
model of each element of $\Sigma$; $A$ and $B$ are {\em equivalent} 
(written $A \equiv_4 B$) iff they have the same models; $\Sigma$ {\em 
entails} $B$ (written $\Sigma \entailfv B$) iff all models of $\Sigma$ 
are models of $B$.  Without loss of generality, we can restrict our 
attention to propositions in NNF only, as $\notf \notf A \equiv_4 A$, 
$\notf (A \andf B) \equiv_4 \notf A \orf \notf B$ and $\notf (A \orf 
B) \equiv_4 \notf A \andf \notf B$ hold.  For easy of notation, we 
will write $A \entailfv B$ in place of $\{ A \} \entailfv B$.

The following relations can easily be verified:

\begin{center}
\begin{tabular}{lcl}
$A \andf B \entailfv A$ \\
$A_{1} \entailfv A_{2}$ and $A_{2} \entailfv A_{3}$ implies $A_{1} \entailfv 
A_{3}$ \\
$A \entailfv A \orf B$ \\
$A \andf (\notf A \orf B) \not \entailfv B$ \\
$A \entailfv B$ implies $\notf B \entailfv \notf A$ \\ 
$A \entailfv B$ implies $A \entailtv B$ .
\end{tabular}
\end{center}

\nd Note that there are no tautologies, \ie \ there is no $A$ such 
that $\entailfv A$, \eg\ $\not\entailfv p \orf \notf p$ (consider $\I$ 
such that $\highi{p} = \emptyset$).  Moreover, every theory is 
satisfiable.  Hence, $p \andf \notf p \not \entailfv q$, as there is a 
model ${\I}$ ($\highi{p} = \{t,f\}$, $\highi{q} = \emptyset$) of $p 
\andf \notf p$ not satisfying $q$.  Moreover, $\entailfv$ is a subset 
of classical entailment $\entailtv$, \ie\ $\entailfv$ is sound w.r.t.\ classical 
entailment.

In \cite{Straccia97a} a simple procedure, deciding whether $\Sigma 
\entailfv A$ holds, has been presented.  The calculus, a tableaux, is 
based on {\em signed propositions of type $\alpha$} (``conjunctive 
propositions'') and of {\em type $\beta$} (``disjunctive 
propositions'') and on their {\em components} which are defined as 
usual \cite{Smullyan68}:\footnote{$\T$ and $\NT$ play the role of 
``$\T$rue" and ``{\tt N}ot $\T$rue", respectively.  In classical 
calculi $\NT$ may be replaced with {\tt F} (``{\tt F}alse").}

\begin{center}
{\footnotesize
\begin{tabular}{cc}
\begin{tabular}{|c|c|c|} \hline
$\alpha$ & $\alpha_1$ & $\alpha_2$ \\ \hline
$\T A \andf B $ & $\T A$ & $\T B$ \\
$\NT A \orf B$ & $\NT A$ & $\NT B$ \\ \hline
\end{tabular} &
\begin{tabular}{|c|c|c|} \hline
$\beta$ & $\beta_1$ & $\beta_2$ \\ \hline
$\T A \orf B$ & $ \T A$ & $ \T B$ \\
$\NT  A \andf B$ & $\NT  A$ & $\NT  B$ \\ \hline
\end{tabular}
\end{tabular}
}
\end{center}

\nd $\T A$ and $\NT A$ are called {\em conjugated signed 
propositions}.  An interpretation $\I$ {\em satisfies} $\T A$ iff $\I$ 
satisfies $A$, whereas $\I$ {\em satisfies} $\NT A$ iff $\I$ does not 
satisfy $A$.  A set of signed propositions is {\em satisfiable} iff 
each element of it is satisfiable.  Therefore,

\begin{equation} \label{fvsat}
\Sigma \entailfv A \mbox{ iff } \T \Sigma \cup \{\NT A\} \mbox{ is not satisfiable}, 
\end{equation}

\nd where $\T \Sigma = $ $\{\T A : A \in \Sigma\}$.  

We present here a simplified version of the calculus for signed 
propositions in NNF, which is based on the set of rules, 
${\mathcal R}^{T}_{4} = \{(\bottomf^{4}), (\andf^{4}), (\orf^{4})\}$, 
described in Table~\ref{calculusfv}.
\begin{table}
    \caption{Simple Tableaux inference rules for four-valued \langL.} \label{calculusfv}
\begin{center}    
\begin{tabular}{l}
\begin{tabular}{cl} \\
 $\ruleal{$(\bottomf^{4})$}
{\T p, \NT p} 
{\bottomf}$ & \\  
    
$\ruleal{$(\andf^{4})$}
{\alpha} 
{\alpha_1, \alpha_2}$ & \\ \\
$\rulebl{$(\orf^{4})$}
{\beta} 
{\beta_{1}}
{\beta_{2}}$  & \\ 
\end{tabular}
\end{tabular}
\end{center}
\end{table}
With ${\sf SAT}_{4}$ we indicate the decision procedure 
that decides whether a set of signed propositions is (four-valued) 
satisfiable or not: ${\sf SAT}_{4}$ derives directly from ${\sf SAT}$ 
in Table~\ref{tabsat}, where the deduction rules ${\mathcal R}^{T}$ 
for \langLf\ have been replaced with the set of rules ${\mathcal 
R}^{T}_{4}$ for four-valued propositional logic.

It has been shown in \cite{Straccia97a} that $\Sigma \entailfv A$ iff 
${\sf SAT}_{4}(\T \Sigma \cup \{ \NT A\})$ returns false.  For 
instance, Figure~\ref{fig2} is a closed deduction tree for $p \andf (q 
\orf r) \entailfv (p \orf r) \andf (q \orf r \orf s)$.
\begin{figure}
\begin{center}
\begin{tabular}{c}
\setlength{\unitlength}{.5cm}
\begin{picture}(16,13)
   \put(1,11){\makebox(14,1){$\T p \andf (q \orf r), 
                             \NT (p \orf r) \andf (q \orf r \orf s)$}}
   \put(7,11){\line(0,-1){1}}

   \put(4,9){\makebox(6,1){$\T p, \T q \orf r$}}
   \put(7,9){\line(-5,-1){5}}
   \put(7,9){\line(5,-1){5}}
   \put(0,7){\makebox(4,1){$\NT p \orf r$}}
   \put(2,7){\line(0,-1){1}}
   \put(0,5){\makebox(4,1){$\NT p, \NT r$}}
   \put(2,5){\line(0,-1){1}}
   \put(0,3){\makebox(4,1){$\bottomf$}}   
   \put(1,2){\makebox(2,1){$\times$}}
   \put(9,7){\makebox(6,1){$\NT q \orf r \orf s$}}
   \put(12,7){\line(0,-1){1}}   
   \put(10,5){\makebox(4,1){$\NT q, \NT r, \NT s$}}
   \put(12,5){\line(-5,-1){3}}
   \put(12,5){\line(5,-1){3}}
   \put(6,3){\makebox(6,1){$\T q$}}
   \put(9,3){\line(0,-1){1}}      
   \put(7,1){\makebox(4,1){$\bottomf$}}   
   \put(8,0){\makebox(2,1){$\times$}}   
   \put(12,3){\makebox(6,1){$\T r$}}
   \put(15,3){\line(0,-1){1}}      
   \put(13,1){\makebox(4,1){$\bottomf$}}   
   \put(14,0){\makebox(2,1){$\times$}}
\end{picture} 
\end{tabular}
\end{center}
\caption{Deduction tree for {\small $p \andf (q \orf r) \entailfv (p \orf r) \andf (q \orf
r \orf s)$}.} \label{fig2}
\end{figure}
Note that, if we switch to the classical two-valued setting, 
soundness and completeness is obtained by extending signed 
propositions as usual: just consider additionally the 
following signed propositions of type $\alpha$.

\begin{center}
\begin{tabular}{|c|c|c|} \hline
$\alpha$ & $\alpha_1$ & $\alpha_2$ \\ \hline
$\T \notf A$ & $\NT  A$ & $\NT  A$ \\
$\NT \notf A$ & $\T  A$ & $\T  A$ \\ \hline
\end{tabular} 
\end{center}

\nd Therefore, in the general case the only difference between 
four-valued and two-valued semantics relies on the negation 
connective.  This is not a surprise as we already said that the 
semantics for the negation is constructive, \ie~expressed in terms of 
$\in$ rather than on $\not \in$. 

The following proposition establishes correctness and completeness of 
the ${\sf SAT}_{4}$ procedure.

\begin{proposition}[\cite{Straccia97a}] \label{satcompletefv}
Let $S$ be a set of signed propositions.  Then ${\sf SAT}_{4}(S)$ 
iff $S$ is four-valued  satisfiable.  
\end{proposition}

\section{Relations among fuzzy entailment and four-valued 
entailment } \label{therelation}

\nd The objective of this section is to establish some relationships 
between fuzzy and four-valued propositional logic.

At first, we show that

\begin{proposition}\label{p6}
Let $A$ and $B$ be two propositions.  If  \ii{i} $A \entailfv B$ or 
\ii{ii} $\entailtv \notf A \andf B$ then for all $n> 0$, $\fuzzy{A}{\geq}{n} 
\entailf \fuzzy{B}{\geq}{n}$.  
\end{proposition}

\begin{proof}
\ii{i} Assume $A \entailfv B$ and suppose to the contrary that there is an 
$n'>0$ such that $\fuzzy{A}{\geq}{n'} \notentailf 
\fuzzy{B}{\geq}{n'}$.  Therefore, there is a fuzzy interpretation $\I'$ 
such that $\highv{A}{\I'} \geq n'$ and $\highv{B}{\I'} < n'$. Let $\I$ be the 
following four-valued interpretation:

\[
\begin{array}{lcl}
\truef{p} & \miff & \highv{p}{\I'} \geq n' \\
\falsef{p} & \miff & 1- \highv{p}{\I'} \geq n' .
\end{array}
\]

\nd We show on induction on the structure of a proposition $C$ that
$\I'$ satisfies $\fuzzy{C}{\geq}{n'}$ iff $\truef{C}$.

\begin{description}
\item[Case letter $p$] If $\I'$ satisfies $\fuzzy{p}{\geq}{n'}$ then
$\highv{p}{\I'}\geq n'$. By definition, $\truef{p}$ follows.
If $\I'$ does not satisfy $\fuzzy{p}{\geq}{n'}$ then
$\highv{p}{\I'} < n'$. By definition, $\nottruef{p}$ follows.

\item[Case literal $\notf p$] If $\I'$ satisfies $\fuzzy{\notf 
p}{\geq}{n'}$ then $1- \highv{p}{\I'}\geq n'$.  By definition, 
$\falsef{p}$ follows and, thus, $\truef{\notf p}$.  If $\I'$ does not 
satisfy $\fuzzy{\notf p}{\geq}{n'}$ then $1- \highv{p}{\I'} < n'$.  By 
definition, $\notfalsef{p}$ follows and, thus, $\nottruef{\notf p}$.

\item[Case $A_{1} \andf A_{2}$] If $\I'$ satisfies 
$\fuzzy{A_{1} \andf A_{2}}{\geq}{n'}$ then $\highv{A_{1}}{\I'}\geq n'$ 
and $\highv{A_{2}}{\I'}\geq n'$.  By induction on $A_{1}$ and $A_{2}$, 
both $\truef{A_{1}}$ and $\truef{A_{2}}$ hold and, thus, 
$\truef{(A_{1} \andf A_{2})}$.

\nd
The case  $A_{1} \orf A_{2}$ is similar.
\end{description}

\nd As a consequence, since $\I'$ satisfies $\fuzzy{A}{\geq}{n'}$ but 
not $\fuzzy{B}{\geq}{n'}$, it follows that $\truef{A}$ and 
$\nottruef{B}$, which is contrary to the assumption $A \entailfv B$. 

\nd \ii{ii} Assume that $\entailtv \notf A \andf B$ holds, \ie\ 
$\entailtv B$ and $\entailtv \notf A$.  Consider $n \in (0,1]$.  
Then either $n \leq 0.5$ or $n > 0.5$.  From $\entailtv B$ it follows 
that $glb(\emptyset, B) = 0.5$, \ie\ $\entailf \fuzzy{B}{\geq}{n}$.  
Therefore, for $n \leq 0.5$, $\fuzzy{A}{\geq}{n} \entailf 
\fuzzy{B}{\geq}{n}$ follows.  From $\entailtv \notf A$, 
$glb(\emptyset, \notf A) = 0.5$ follows, \ie\ $lub(\emptyset, A) = 
0.5$.  As a consequence, for $n > 0.5$, $\fuzzy{A}{\geq}{n}$ is 
unsatisfiable and, thus, $\fuzzy{A}{\geq}{n} \entailf 
\fuzzy{B}{\geq}{n}$ holds.  Therefore, for all $n> 0$ 
$\fuzzy{A}{\geq}{n} \entailf \fuzzy{B}{\geq}{n}$ holds.  
\end{proof}

\begin{proposition}\label{p7}
Let $A$ and $B$ be two propositions.  If $\not\entailtv B$ and $A 
\not\entailfv B$ then there is a four-valued interpretation $\I$ such 
that $\truef{A}$, $\nottruef{B}$ and for no letter $p$, $\highi{p} = 
\emptyset$.
\end{proposition}

\begin{proof}
Since $A \not\entailfv B$, ${\sf SAT}_{4}(\{\T A, \NT B\})$ returns 
true.  Therefore, there is a completed branch $\phi$.  Suppose that 
for each completed branch $\phi_{i}$, $1\leq i \leq b$, there is a 
letter $p_{i}$ occurring in $B$ such that both $\NT p_{i} \in 
\phi_{i}^{M}$ and $\NT \notf p_{i} \in \phi_{i}^{M}$.  As a 
consequence, collecting all the $\NT$ expressions in branches 
$\phi_{i}$, informally $\NT B$ is equivalent to

\[
\begin{array}{lcl}
\NT B &\equiv& \bigvee_{i=1}^{b} (\NT p_{i} \andf \NT\notf p_{i} \andf \NT F_{i})   \\ 
      & \equiv &  \bigvee_{i=1}^{b} (\NT (p_{i} \orf \notf p_{i} \orf F_{i})        \\
      & \equiv & \NT\bigwedge_{i=1}^{b} (p_{i} \orf \notf p_{i} \orf F_{i})
\end{array}
\]

\nd and, thus, $B$ is classically equivalent to $\bigwedge_{i} (p_{i} 
\orf \notf p_{i} \orf F_{i})$. It follows that $\entailtv B$, contrary 
to our assumption. Therefore, there is  a \emph{completed} branch $\phi$ such 
that for any letter $p$ occurring in $B$, 

\begin{enumerate}
    \item if $\NT p \in \phi^{M}$ 
    then $\NT \notf p \not\in \phi^{M}$ and $\T p \not\in \phi^{M}$;
    
    \item if $\NT \notf p \in \phi^{M}$ then $\NT p \not\in \phi^{M}$ 
    and $\T \notf p \not\in \phi^{M}$.
    
\end{enumerate} 

\nd
Let $\I$ be the following four-valued interpretation:

\[
\begin{array}{lcl}
\truef{p} & \miff & \T p \in \phi^{M} \\
\falsef{p} & \miff & \T \notf p \in \phi^{M} \\
\highi{p} =\{f\} & \miff & \NT p \in \phi^{M} \\
\highi{p} =\{t\} & \miff & \NT \notf p \in \phi^{M} .
\end{array}
\]

\nd It follows that for any letter $p$, $\highi{p} \neq \emptyset$.  
Furthermore, it can easily be shown on induction on the structure of 
$A$ and $B$ that $\I$ satisfies both $\T A$ and $\NT B$.  Therefore, 
$\truef{A}$ and $\nottruef{B}$.
\end{proof}

\begin{proposition}\label{p8}
Let $A$ and $B$ be two propositions and consider $0 < n\leq 0.5$.  
$\fuzzy{A}{\geq}{n} \entailf \fuzzy{B}{\geq}{n}$ iff $\entailtv B$ or 
$A \entailfv B$.
\end{proposition}

\begin{proof} \mbox{ \ } \\
$\Rightarrow.)$ Assume $0< n \leq 0.5$ and $\fuzzy{A}{\geq}{n} \entailf 
\fuzzy{B}{\geq}{n}$.  If $\entailf \fuzzy{B}{\geq}{n}$ then $\entailtv 
B$, by Proposition~\ref{prop1}.  Otherwise, $\notentailf 
\fuzzy{B}{\geq}{n}$ implies $\not\entailtv B$ (as a NNF of $\notf 
\fuzzy{B}{\geq}{n}$ is normalised and by Proposition~\ref{prop2}).  
So, let us show that $A \entailfv B$.  Suppose to the contrary that $A 
\not\entailfv B$.  From Proposition~\ref{p7}, there is an 
interpretation $\I$ such that $\truef{A}$, $\nottruef{B}$ and for no 
letter $p$, $\highi{p} = \emptyset$. Consider the following fuzzy 
interpretation $\I'$:

\begin{enumerate}
    \item if $\highi{p} = \{t\}$ then $\highv{p}{\I'} = 1$;
    \item if $\highi{p} = \{f\}$ then $\highv{p}{\I'} = 0$;
    \item if $\highi{p} = \{t,f\}$ then $\highv{p}{\I'} = 0.5$.   
\end{enumerate}    

\nd Let us show on induction of the structure of any proposition $C$ 
and any $0<n \leq 0.5$ that $\truef{C}$ iff $\highv{C}{\I'} \geq n$ holds.

\begin{description}
\item[Case letter $p$] By definition, $\truef{p}$ implies 
$\highv{p}{\I'} = 1$ and, thus, $\highv{p}{\I'} \geq n$.  On the other 
hand, $\nottruef{p}$ implies $\highi{p} = \{f\}$ and, thus, $\highv{p}{\I'} = 
0$.  As a consequence, $\highv{p}{\I'} < n$;

\item[Case literal $\notf p$]  $\truef{(\notf p)}$ implies $\falsef{p}$. 
Therefore, either   $\highv{p}{\I'} = 0$ or $\highv{p}{\I'} = 0.5$. As 
a consequence, $\highv{(\notf p)}{\I'}= 1- \highv{p}{\I'} \geq n$ ($n \leq 
0.5$). On the other hand, $\nottruef{(\notf p)}$ implies 
$\notfalsef{p}$. Therefore, $\highi{p} = \{t\}$ and, by definition, 
$\highv{p}{\I'} = 1$ follows. As a consequence, $\highv{(\notf p)}{\I'}= 
1- \highv{p}{\I'} = 0 < n$;

\item[Cases $A_{1} \andf A_{2}$ and $A_{1} \orf A_{2}$] Straightforward. 
\end{description}

\nd Therefore, from $\truef{A}$ it follows that $\I'$ satisfies 
$\fuzzy{A}{\geq}{n}$.  From $\nottruef{B}$ it follows that $\I'$ does 
not satisfy $\fuzzy{B}{\geq}{n}$, contrary to the assumption that 
$\fuzzy{A}{\geq}{n} \entailf \fuzzy{B}{\geq}{n}$ holds. 

$\Leftarrow.)$ From $A \entailfv B$ and from Proposition~\ref{p6} for 
all $n \in (0,1]$, $\fuzzy{A}{\geq}{n} \entailf \fuzzy{B}{\geq}{n}$ 
follows.  Otherwise, if $\entailtv B$ then $\entailf 
\fuzzy{B}{\geq}{0.5}$ and, thus, for any $0 < n\leq 0.5$ 
$\fuzzy{A}{\geq}{n} \entailf \fuzzy{B}{\geq}{n}$.  
\end{proof}

\nd Proposition~\ref{p8} can be generalised as follows.  At first, we 
show that

\begin{proposition}\label{p8a}
Let $\psi_{1}$ and $\psi_{2}$ be two meta propositions such that 
$\psi_{1}$ is sub-normalised and let $n \in (0,0.5]$.  If $\psi_{1} 
\entailf \psi_{2}$ then $\fuzzy{\crisp{\psi_{1}}}{\geq}{n} 
\entailf \fuzzy{\crisp{\psi_{2}}}{\geq}{n}$. 
\end{proposition}

\begin{proof}
Assume $\psi_{1} \entailf \psi_{2}$.  Mark all meta-literals in 
$\psi_{2}$ with $^{*}$.  Consider a deduction of ${\sf 
SAT}(\{\psi_{1}, \notf \psi_{2}\})$, which returns false, and let $T$ 
be the deduction tree.  As a consequence, all branches $\phi$ in $T$ 
are closed.

Let us consider the following substitution, $\mytrans{(\cdot)}$, in each 
branch $\phi$.  For each meta literal $\psi$ occurring in $\phi^{M}$, 
\ii{i} if $\psi =\fuzzy{p}{\ r \ }{m}$ is not marked with $^{*}$ then 
for $r \in \{\geq,>\}$ replace $\psi$ with $\fuzzy{p}{\ r \ }{n}$ and 
for $r \in \{\leq,<\}$ replace $\psi$ with $\fuzzy{p}{\ r \ }{1-n}$; 
and \ii{ii} if $\psi =\fuzzy{p}{\ r \ }{m}^{*}$ is marked with $^{*}$ 
then for $r \in \{\geq,>\}$ replace $\psi$ with $\fuzzy{p}{>}{1-n}$ 
and for $r \in \{\leq,<\}$ replace $\psi$ with $\fuzzy{p}{<}{n}$.  
$\bottomf$ is mapped into it.  Let $\mytrans{\psi}$ and 
$\mytrans{\phi}$ be the result of this substitution, for each meta 
proposition $\psi$ and for each (closed) branch $\phi$, respectively.

We show on induction of the depth $d$ of each branch $\phi$ in the 
deduction tree $T$, that $\mytrans{\phi}$ is a branch in a deduction tree 
of ${\sf SAT}(\{\fuzzy{\crisp{\psi_{1}}}{\geq}{n}, 
\fuzzy{\crisp{\psi_{2}}}{<}{n}\})$ and, thus, the tree $\mytrans{T}$ 
formed out by the (closed) branches $\mytrans{\phi}$, for $\phi$ branch 
in $T$, is a closed deduction tree for ${\sf 
SAT}(\{\fuzzy{\crisp{\psi_{1}}}{\geq}{n}, 
\fuzzy{\crisp{\psi_{2}}}{<}{n}\})$.  Therefore, 
$\fuzzy{\crisp{\psi_{1}}}{\geq}{n} \entailf 
\fuzzy{\crisp{\psi_{2}}}{\geq}{n}$.

\begin{description}
\item[Case $d=1$] Therefore, there is an unique closed branch $\phi$ 
in $T$ as the result of the application of the $(\bottomf)$ rule, \ie 
\ $\phi^{M} = \{\psi_{1}, \notf \psi_{2}, \bottomf\}$.  Since $\phi$ 
is closed, $ctd(\psi_{1},\notf\psi_{2})$.  There are eight possible 
cases for $r,r' \in \{\geq,>, \leq,<\}$ such that $\psi_{1} = 
\fuzzy{p}{\ r\ }{k}$, $\notf\psi_{2} \equivf \fuzzy{p}{\ r'\ }{m}^{*}$ 
and $ctd(\psi_{1},\notf\psi_{2})$.  Let us consider the cases \ii{a} 
$\psi_{1} = \fuzzy{p}{\geq}{k}$, $\notf\psi_{2} \equivf 
\fuzzy{p}{\leq}{m}^{*}$.  By definition, $\mytrans{\phi}^{M}$ is 
$\{\fuzzy{p}{\geq}{n}, \fuzzy{p}{<}{n}^{*}, \bottomf\}$.  Therefore, 
$\mytrans{\phi}$ is a closed branch of a deduction tree for ${\sf 
SAT}(\{\fuzzy{\crisp{\psi_{1}}}{\geq}{n}, 
\fuzzy{\crisp{\psi_{2}}}{<}{n}\})$ $=$ ${\sf 
SAT}(\{\fuzzy{p}{\geq}{n}, \fuzzy{p}{<}{n}\})$; \ii{b} $\psi_{1} = 
\fuzzy{p}{\leq}{k}$, $\notf \psi_{2} \equivf \fuzzy{p}{\geq}{m}^{*}$.  
By definition, $\mytrans{\phi}^{M}$ is $\{\fuzzy{p}{\leq}{1-n}, 
\fuzzy{p}{>}{1-n}^{*}, \bottomf\}$.  Therefore, $\mytrans{\phi}$ is a 
closed branch of a deduction tree for ${\sf 
SAT}(\{\fuzzy{\crisp{\psi_{1}}}{\geq}{n}, 
\fuzzy{\crisp{\psi_{2}}}{<}{n}\})= {\sf SAT}(\{\fuzzy{\notf 
p}{\geq}{n}, \fuzzy{\notf p}{<}{n}\})$.  The other cases can be shown 
similarly.

\item[Case $d>1$] Consider a branch $\phi$ of depth $d > 1$.  $\phi$ 
is the result of the application of one of the rules in ${\mathcal 
R}^{T}$ to a branch $\phi'$ of depth $d-1$.  On induction on $\phi'$, 
$\mytrans{\phi'}$ is a branch in a deduction tree of ${\sf 
SAT}(\{\fuzzy{\crisp{\psi_{1}}}{\geq}{n}, 
\fuzzy{\crisp{\psi_{2}}}{<}{n}\})$.  Let us show that $\mytrans{\phi}$ 
is still a branch in a deduction tree of ${\sf 
SAT}(\{\fuzzy{\crisp{\psi_{1}}}{\geq}{n}, 
\fuzzy{\crisp{\psi_{2}}}{<}{n}\})$.  \ii{1} Suppose that rule 
$(\andf)$ has been applied to $\psi \andf \psi' \in \phi'^{M}$ and, 
thus, $\psi, \psi' \in \phi^{M}$.  By definition of 
$\mytrans{(\cdot)}$, $\mytrans{(\psi \andf \psi')}$ is in 
$\mytrans{\phi'}^{M}$, \ie\ $\mytrans{\psi} \andf \mytrans{\psi'}$ is 
in $\mytrans{\phi'}^{M}$.  As a consequence, the $(\andf)$ rule can be 
applied to it and, thus, $\mytrans{\psi}$ and $\mytrans{\psi'}$ are in 
$\mytrans{\phi'}^{M}$.  \ii{2} the case of rule $(\orf)$ is similar.  
Finally, \ii{3} suppose that rule $(\bottomf)$ has been applied to 
literals $\psi, \psi' \in \phi'^{M}$ such that $ctd(\psi, \psi')$ and 
$\bottomf \in \phi^{M}$.  By definition of $\mytrans{(\cdot)}$, 
$\mytrans{\psi}$ and $\mytrans{\psi'}$ are in $\mytrans{\phi'}^{M}$.  
Now, we proceed similarly to the case $d=1$.  As $\psi_{1}$ is 
sub-normalised, either $\psi$ or $\psi'$ has to be marked with $^{*}$.  
Without loss of generality, we can distinguish two cases \ii{i} only 
$\psi'$ is marked with $^{*}$; and \ii{ii} both $\psi$ and $\psi'$ are 
marked with $^{*}$.  Let us consider case \ii{i}.  There are eight 
possible cases for $r,r' \in \{\geq,>, \leq,<\}$ such that $\psi = 
\fuzzy{p}{\ r\ }{n}$, $\psi'= \fuzzy{p}{\ r'\ }{m}^{*}$ and 
$ctd(\psi,\psi')$.  Let us consider the case \ii{a} $\psi = 
\fuzzy{p}{\geq}{k}$, $\psi'= \fuzzy{p}{\leq}{m}^{*}$.  By definition, 
$\mytrans{\phi'}^{M}$ contains both $\fuzzy{p}{\geq}{n}$ and 
$\fuzzy{p}{<}{n}^{*}$, which are pairwise contradictory.  Therefore, 
rule $(\bottomf)$ can be applied to $\mytrans{\phi'}$ and 
$\mytrans{\phi}^{M}$ contains $\bottomf$ and, thus, $\phi$ is closed; 
\ii{b} $\psi = \fuzzy{p}{\leq}{k}$, $\psi'= \fuzzy{p}{\geq}{m}^{*}$.  
By definition, $\mytrans{\phi'}^{M}$ contains both 
$\fuzzy{p}{\leq}{1-n}$ and $\fuzzy{p}{>}{1-n}^{*}$, which are pairwise 
contradictory.  Therefore, rule $(\bottomf)$ can be applied to 
$\mytrans{\phi'}$ and $\mytrans{\phi}^{M}$ contains $\bottomf$ and, 
thus, $\mytrans{\phi}$ is closed.  The other cases are similar.  
Finally, consider the case \ii{ii} both $\psi$ and $\psi'$ are marked 
with $^{*}$.  Without loss of generality, there are four possible 
cases for $r,r' \in \{\geq,>, \leq,<\}$ such that $\psi = \fuzzy{p}{\ 
r\ }{n}^{*}$, $\psi'= \fuzzy{p}{\ r'\ }{m}^{*}$ and $ctd(\psi,\psi')$.  
Let us consider the case $\psi = \fuzzy{p}{\geq}{k}^{*}$, $\psi'= 
\fuzzy{p}{\leq}{m}^{*}$.  By definition, $\mytrans{\phi'}^{M}$ contains 
both $\fuzzy{p}{>}{1-n}$ and $\fuzzy{p}{<}{n}^{*}$, which are pairwise 
contradictory, for $n \in (0,0.5]$. Then proceed similarly as 
above.  The other cases are similar. 
\end{description}
\end{proof}

\nd Note that the converse of the above proposition does not hold.  
For instance, given $n \in (0,0.5]$, $\fuzzy{p}{\geq}{n} \entailf \fuzzy{p 
\orf q}{\geq}{n}$, but $\fuzzy{p}{\geq}{0.2} \notentailf 
\fuzzy{p}{\geq}{0.3} \orf \fuzzy{p}{\geq}{0.1}$.

\begin{proposition}\label{p8b}
Let $\psi_{1}$ and $\psi_{2}$ be two meta propositions such that 
$\psi_{1}$ is sub-normalised. If $\psi_{1} \entailf \psi_{2}$ then either 
$\entailtv \crisp{\psi_{2}}$ or $\crisp{\psi_{1}} \entailfv 
\crisp{\psi_{2}}$.  
\end{proposition}

\begin{proof}
From hypothesis, from Proposition~\ref{p8} and Proposition~\ref{p8a} 
it follows immediately that either $\entailtv \crisp{\psi_{2}}$ or 
$\crisp{\psi_{1}} \entailfv \crisp{\psi_{2}}$ holds.
\end{proof}    

\nd As a meta theory is equivalent to a conjunction of meta 
propositions, we have immediately,

\begin{proposition}\label{p8bb}
Let $\FKB$ be a sub-normalised meta theory and let $\psi$ be a meta 
proposition.  If $\FKB \entailf \psi$ then either $\entailtv 
\crisp{\psi}$ or $\crisp{\FKB} \entailfv \crisp{\psi}$.  

\end{proposition}

\nd The converse of the above propositions does not hold.  Indeed, $p 
\entailfv p \orf q$, but $\fuzzy{p}{\geq}{0.2} \notentailf 
\fuzzy{p}{\geq}{0.3} \orf \fuzzy{p}{\geq}{0.1}$.

Dually to Proposition~\ref{p8} we have

\begin{proposition}\label{p8c}
Let $A$ and $B$ be two propositions and consider $0.5 < n\leq 1$.  
$\fuzzy{A}{\geq}{n} \entailf \fuzzy{B}{\geq}{n}$ iff for a DNF $A_{1} 
\orf \ldots \orf A_{l}$ of $A$ and for each $j=1, \ldots l$, either 
\ii{i} $\entailtv \notf A_{j}$ or \ii{ii} $A_{j} \entailfv B$.

\end{proposition}

\begin{proof} \mbox{ \ } \\
$\Rightarrow.)$ Assume $n > 0.5$ and $\fuzzy{A}{\geq}{n} \entailf 
\fuzzy{B}{\geq}{n}$.  Consider a DNF $A_{1}\orf \ldots \orf A_{l}$ 
of $A$.  From $\fuzzy{A}{\geq}{n} \entailf \fuzzy{B}{\geq}{n}$ it 
follows that $\bigvee_{j=1}^{l}\fuzzy{A_{j}}{\geq}{n} \entailf 
\fuzzy{B}{\geq}{n}$ and, thus, for each $j=1, \ldots l$, 
$\fuzzy{A_{j}}{\geq}{n} \entailf \fuzzy{B}{\geq}{n}$, \ie\ $S_{j} = 
\{\fuzzy{A_{j}}{\geq}{n}, \fuzzy{B}{<}{n} \}$ is unsatisfiable.  Mark 
all the meta literals in a NNF of $B$ with $^{*}$.  Let $\phi_{1}, 
\ldots, \phi_{h}$ be all the branches of a deduction of ${\sf 
SAT}(S_{j})$.  Consider a branch $\phi_{i}$.  Obviously, $\phi_{i}$ is 
closed.  Therefore, there are meta literals $\psi_{i}, \psi_{i}' \in 
\phi_{i}^{M}$ such that $ctd(\psi_{i},\psi_{i}')$.  Consider the four 
pairs for $\psi_{i}$ and $\psi'_{i}$, respectively:

\begin{eqnarray}
\fuzzy{p}{\geq}{n} & , & \fuzzy{p}{\leq}{1-n}  \label{ctd1}\\
\fuzzy{p}{\geq}{n} &  , & \fuzzy{p}{<}{n}^{*}  \label{ctd2} \\
\fuzzy{p}{\leq}{1-n} & ,  & \fuzzy{p}{>}{1-n}^{*}  \label{ctd3} \\
\fuzzy{p}{<}{n}^{*} &  ,  & \fuzzy{p}{>}{1-n}^{*}  \label{ctd4}  
\end{eqnarray}

\nd At first, if \myref{ctd1} is the case ($n>0.5$) then $A_{j}$ is 
unsatisfiable, \ie\ $\entailtv \notf A_{j}$ and, thus, condition 
\ii{i} is satisfied. Second, \myref{ctd4} cannot be the case as $n > 
0.5$. So, for the other cases, we can assume that $A_{j}$ is satisfiable.

Let us consider the following transformation $\myStrans{\cdot}$ for 
each branch $\phi_{i}$.  For each meta literal $\psi$ occurring in 
$\phi^{M}$, \ii{i} if $\psi =\fuzzy{p}{\geq}{n}$ is not marked with 
$^{*}$ then $\psi \mapsto \T p$; \ii{ii} if $\psi =\fuzzy{p}{\leq}{1-n}$ is not 
marked with $^{*}$ then $\psi \mapsto \T \notf p$; \ii{iii} if $\psi 
=\fuzzy{p}{<}{n}^{*}$ is marked with $^{*}$ then $\psi \mapsto \NT p$; 
and \ii{iv} if $\psi =\fuzzy{p}{>}{1-n}^{*}$ is marked with $^{*}$ 
then $\psi \mapsto \NT \notf p$.  Let $\myStrans{\psi}$, 
$\myStrans{\phi}$ and $\myStrans{S}$  be the result of this transformation, for each meta 
proposition $\psi$, for each branch $\phi_{i}$ and for each set of 
meta propositions $S$, respectively.

Similarly to Proposition~\ref{p8a}, it can be shown on induction of 
the depth $d$ of each branch $\phi_{i}$ of a deduction ${\sf SAT}(S_{j})$, 
that the branch $\myStrans{\phi_{i}}$ is a closed branch of a 
four-valued deduction ${\sf SAT}_{4}(\myStrans{S_{j}})$.  But, 
$\myStrans{S_{j}}$ is $\{\T A_{j}, \NT B\}$ and, thus, $A_{j} \entailfv 
B$.  In the induction proof, it suffices to show that if 
$ctd(\psi_{i},\psi_{i}')$ then 
$ctd(\myStrans{\psi_{i}},\myStrans{\psi_{i}'})$, \ie\ if the 
$(\bottomf)$ rule is applicable to $\phi_{i}$ then the $(\bottomf)^{4}$ 
rule is applicable to $\myStrans{\phi_{i}}$.  For the other rules the 
proof is immediate.  So, as we have seen above, either case 
\myref{ctd2} or case \myref{ctd3} holds.  If \myref{ctd2} is the case 
then $\myStrans{\psi} = \T p$ and $\myStrans{\psi'} = \NT p$ and, 
thus, $ctd(\myStrans{\psi_{i}},\myStrans{\psi_{i}'})$.  Otherwise, if 
\myref{ctd3} is the case then $\myStrans{\psi} = \T \notf p$ and 
$\myStrans{\psi'} = \NT \notf p$ and, thus, 
$ctd(\myStrans{\psi_{i}},\myStrans{\psi_{i}'})$, which completes 
$\Rightarrow.)$.

$\Leftarrow.)$ Consider $n > 0.5$.  It suffices to show that for each 
$j=1, \ldots l$, if either \mbox{$\entailtv\notf A_{j}$} or $A_{j} \entailfv 
B$ then $\fuzzy{A_{j}}{\geq}{n} \entailf \fuzzy{B}{\geq}{n}$.  If 
$\entailtv \notf A_{j}$ then $\fuzzy{A_{j}}{\geq}{n}$ is 
unsatisfiable, as $n > 0.5$ and, thus $\fuzzy{A_{j}}{\geq}{n} \entailf 
\fuzzy{B}{\geq}{n}$.  Otherwise, if $A_{j} \entailfv B$ then, by 
Proposition~\ref{p6}, it follows that $\fuzzy{A_{j}}{\geq}{n} \entailf 
\fuzzy{B}{\geq}{n}$.  
\end{proof}

\nd
We conclude with

\begin{proposition}\label{p9}
Let $A$ and $B$ be two propositions, $n_{1} \leq 0.5$ and $n_{2} 
>0.5$.  It follows that, for both $n \in \{n_{1},n_{2}\}$, 
$\fuzzy{A}{\geq}{n} \entailf \fuzzy{B}{\geq}{n}$ iff either \ii{i} $A 
\entailfv B$; or \ii{ii} $\entailtv \notf A \andf B$ holds.  
\end{proposition}

\begin{proof}
	
$\Rightarrow.)$ Assume that for both $n \in \{n_{1},n_{2}\}$, 
$\fuzzy{A}{\geq} {n} \entailf \fuzzy{B}{\geq}{n}$ holds.  If $A 
\entailfv B$ then condition \ii{i} is trivially satisfied.  Otherwise, 
assume $A \not\entailfv B$.  From Proposition~\ref{p8}, $\entailtv B$ 
follows.  But then, we know that $glb(\emptyset, B) = 0.5$.  As a 
consequence, for $n=n_{2} > 0.5$ no interpretation satisfies 
$\fuzzy{B}{\geq}{n}$.  Therefore, since by hypothesis 
$\fuzzy{A}{\geq}{n} \entailf \fuzzy{B}{\geq}{n}$ holds for $n > 0.5$, 
it follows that for $n > 0.5$, $\fuzzy{A}{\geq}{n}$ has to be 
unsatisfiable, \ie\ $\entailf \fuzzy{A}{<}{n}$ and, thus, $\entailf 
\fuzzy{\notf A}{>}{1-n}$, for $n > 0.5$.  As a NNF of $\notf 
\fuzzy{\notf A}{>}{1-n}$ is normalised, from Proposition~\ref{prop2} 
it follows that $\entailtv \notf A$.  Therefore, condition \ii{ii} is 
satisfied.\footnote{An example of case \ii{ii} is the 
following: for $n = n_{1},n_{2}$, where $n_{1} \leq 0.5$ and $n_{2} 
>0.5$, $\fuzzy{p\andf \notf p}{\geq}{n} \entailf \fuzzy{q \orf \notf 
q}{\geq}{n}.$}

$\Leftarrow.)$ From Proposition~\ref{p6} it follows directly that for 
all $n> 0$, $\fuzzy{A}{\geq}{n} \entailf \fuzzy{B}{\geq}{n}$.  In 
particular, $\fuzzy{A}{\geq}{n} \entailf \fuzzy{B}{\geq}{n}$ holds for 
$n = n_{1},n_{2}$, where $n_{1} \leq 0.5$ and $n_{2} >0.5$.  
\end{proof}

\nd An interesting application of the above proposition is the 
following.  Consider the quite natural and common fuzzy entailment 
relation, $\entailf_{r}$, among propositions, defined as follows (see, \eg\ 
\cite{Xiaochun95,Yager85}):

\[
\begin{array}{lcc}
A \entailf_{r} B & \miff & \mbox{ for all fuzzy interpretations $\I$, 
$\highi{A} \leq \highi{B}$}.
\end{array}
\]

\nd
Now, it is quite easy to show that

\begin{proposition}\label{p10}
Let $A$ and $B$ be two propositions.  It follows that $A \entailf_{r} 
B$ iff for all $n> 0$, $\fuzzy{A}{\geq}{n} \entailf_{r} \fuzzy{B}{\geq}{n}$.  
\end{proposition}

\begin{proof}
$\Rightarrow.)$ Assume that $A \entailf_{r} B$.  Suppose to the 
contrary that $\exists n > 0$ such that $\fuzzy{A}{\geq}{n} 
\not\entailf \fuzzy{B}{\geq}{n}$.  Therefore, there is a fuzzy 
interpretation $\I$ such that $\highi{A} \geq n$ and 
$\highi{B} < n$.  But, from the hypothesis $n \leq 
\highi{A} \leq \highi{B} < n$ follows.  Absurd.

$\Leftarrow.)$ Assume that for all $n> 0$, $\fuzzy{A}{\geq}{n} 
\entailf \fuzzy{B}{\geq}{n}$.  Suppose to the contrary that $A 
\not\entailf_{r} B$.  Therefore, there is a fuzzy interpretation $\I$ such 
that $\highi{A} > \highi{B}$.  Consider $\overline{n} = \highi{A}$.  
Of course, $\I$ satisfies $\fuzzy{A}{\geq}{\overline{n}}$.  Therefore, 
from the hypothesis it follows that $\I$ satisfies 
$\fuzzy{B}{\geq}{\overline{n}}$, \ie\ $\highi{B} \geq \overline{n} = 
\highi{A} > \highi{B}$.  Absurd.  
\end{proof}

\nd Finally, we can apply Proposition~\ref{p9} and 
Proposition~\ref{p6} and obtain

\begin{corollary}\label{p11}
Let $A$ and $B$ be two propositions.  It follows that $A  \entailf_{r} 
B$ iff either \ii{i} $A \entailfv B$; or \ii{ii} $\entailtv \notf A 
\andf B$ holds.  
\end{corollary}

\nd Essentially, Corollary~\ref{p11} establishes that \emph{for all 
interesting cases, \ie\ the theory $A$ is classically satisfiable and 
the conclusion B is not a classical tautology, fuzzy entailment 
$ \entailf_{r}$ is equivalent to four-valued entailment $\entailfv$}.  In 
particular, \citeN{Yager85} further restricts $ \entailf_{r}$ to the 
case where the premise should be classically satisfiable and, thus, 
from Corollary~\ref{p11} it follows that $A  \entailf_{r} B$ iff $A 
\entailfv B$.  In fact, a closer look to the axiomatization provided 
by Yager reveals that it is a (not minimal) 
axiomatization for four-valued logic.

Finally, some alternative, still popular, definitions of fuzzy 
entailment are (see, \eg\ \cite{Xiaochun95}):

\begin{enumerate}
\item
$A \entailf_{a} B$ iff  for all fuzzy interpretations $\I$, 
$\max\{1-\highi{A}, \highi{B}\} \geq 0.5$;
\item
$A \entailf_{b} B$ iff  for all fuzzy interpretations $\I$, 
$\highi{A} \geq 0.5$ implies $\highi{B} \geq 0.5$;
\item
$A \entailf_{c} B$ iff for all fuzzy interpretations $\I$, 
$\highi{A} > 0.5$ implies $\highi{B} > 0.5$.
\end{enumerate}

\nd The following relations are easily verified.  

\begin{enumerate}
\item $A \entailf_{a} B$ iff $\entailf \fuzzy{\notf A \orf 
B}{\geq}{0.5}$.  As $\fuzzy{\notf A \orf B}{\geq}{0.5}$ is normalised, 
we already know that this is equivalent to $\entailtv \notf A \orf B$, 
\ie\ $A \entailtv B$.  Therefore, $A \entailf_{a} B$ iff $A \entailtv 
B$.  This result already has been proven differently in \cite{Lee72};

\item $A \entailf_{b} B$ iff $\fuzzy{A}{\geq}{0.5} \entailf 
\fuzzy{B}{\geq}{0.5}$.  From Proposition~\ref{p6} and 
Proposition~\ref{p8} it follows that $A \entailf_{b} B$ iff either 
$\entailtv B$ or $A \entailfv B$;

\item
$A \entailf_{c} B$ iff $\fuzzy{A}{>}{0.5} \entailf \fuzzy{B}{>}{0.5}$.  
According to Proposition~\ref{p8c} $A \entailf_{c} B$ iff for a DNF 
$A_{1} \orf \ldots \orf A_{l}$ of $A$ and for each $j=1, \ldots l$, 
either $\entailtv \notf A_{j}$ or $A_{j} \entailfv B$.

\end{enumerate}

\section{Conclusions} \label{concl}

\nd In this paper we have shown that there is a strict relation 
between various common definitions of fuzzy entailment 
($\entailf_{(\cdot)}$), four-valued entailment ($\entailfv$) and 
two-valued entailment ($\entailtv$).  While the presented results allow to 
describe \emph{qualitatively} what is inferable according to 
$\entailf_{(\cdot)}$, neither $\entailfv$ nor $\entailtv$ can solve 
the \emph{quantitative} aspect, \eg\ the computation of the greatest 
lower bound, $glb(\Sigma,A)$.



%
%



\begin{thebibliography}{}

\bibitem[\protect\citeauthoryear{Anderson and Belnap}{Anderson and
  Belnap}{1975}]{Anderson75}
{\sc Anderson, A.~R.} {\sc and} {\sc Belnap, N.~D.} 1975.
\newblock {\em Entailment - the logic of relevance and necessity}.
\newblock Princeton University Press, Princeton, NJ.

\bibitem[\protect\citeauthoryear{Belnap}{Belnap}{1977}]{Belnap77a}
{\sc Belnap, N.~D.} 1977.
\newblock A useful four-valued logic.
\newblock In {\em Modern uses of multiple-valued logic}, {G.~Epstein} {and}
  {J.~M. Dunn}, Eds. Reidel, Dordrecht, NL, 5--37.

\bibitem[\protect\citeauthoryear{Chen and Kundu}{Chen and
  Kundu}{1996}]{ChenJ96}
{\sc Chen, J.} {\sc and} {\sc Kundu, S.} 1996.
\newblock A sound and complete fuzzy logic system using {Z}adeh's implication
  operator.
\newblock In {\em Proc.\ of the 9th Int.\ Sym.\ on Methodologies for
  Intelligent Systems (ISMIS-96)}, {Z.~W. Ras} {and} {M.~Maciek}, Eds. Number
  1079 in Lecture Notes in Artificial Intelligence. Springer-Verlag, 233--242.

\bibitem[\protect\citeauthoryear{Dunn}{Dunn}{1986}]{Dunn86}
{\sc Dunn, J.~M.} 1986.
\newblock Relevance logic and entailment.
\newblock In {\em Handbook of Philosophical Logic}, {D.~M. Gabbay} {and}
  {F.~Guenthner}, Eds. Vol.~3. Reidel, Dordrecht, NL, 117--224.

\bibitem[\protect\citeauthoryear{Lee}{Lee}{1972}]{Lee72}
{\sc Lee, R. C.~T.} 1972.
\newblock Fuzzy logic and the resolution principle.
\newblock {\em Journal of the ACM\/}~{\em 19,\/}~1 (Jan.), 109--119.

\bibitem[\protect\citeauthoryear{Levesque}{Levesque}{1984}]{Levesque84a}
{\sc Levesque, H.~J.} 1984.
\newblock A logic of implicit and explicit belief.
\newblock In {\em Proc.\ of the 3th Nat.\ Conf.\ on Artificial Intelligence
  (AAAI-84)}. Austin, TX, 198--202.

\bibitem[\protect\citeauthoryear{Pavelka}{Pavelka}{1979}]{Pavelka79}
{\sc Pavelka, J.} 1979.
\newblock On fuzzy logic {I},{II},{III}.
\newblock {\em Zeitschrift f{\"u}r Mathematik und Logik\/}~{\em 25},
  45--52,119--134,447--464.

\bibitem[\protect\citeauthoryear{Smullyan}{Smullyan}{1968}]{Smullyan68}
{\sc Smullyan, R.~M.} 1968.
\newblock {\em First Order Logic}.
\newblock Springer, Berlin.

\bibitem[\protect\citeauthoryear{Straccia}{Straccia}{1997}]{Straccia97a}
{\sc Straccia, U.} 1997.
\newblock A four-valued fuzzy propositional logic.
\newblock In {\em Proc.\ of the 15th Int.\ Joint Conf.\ on Artificial
  Intelligence (IJCAI-97)}. Nagoya, Japan, 128--133.

\bibitem[\protect\citeauthoryear{Straccia}{Straccia}{1999}]{Straccia99a}
{\sc Straccia, U.} 1999.
\newblock Foundations of a logic based approach to multimedia document
  retrieval.
\newblock Ph.D. thesis, Department of Computer Science, University of Dortmund,
  Dortmund, Germany.

\bibitem[\protect\citeauthoryear{Straccia}{Straccia}{2000}]{Straccia00a}
{\sc Straccia, U.} 2000.
\newblock Reasoning and experimenting within {Z}adeh's fuzzy propositional
  logic.
\newblock Technical Report 2000-b4-011, Istituto di Elaborazione
  dell'Informazione, Consiglio Nazionale delle Ricerche, Pisa, Italy.
\newblock Submitted for pubblication.

\bibitem[\protect\citeauthoryear{Xiachun, Yunfei, and Xuhua}{Xiachun
  et~al\mbox{.}}{1995}]{Xiaochun95}
{\sc Xiachun, C.}, {\sc Yunfei, J.}, {\sc and} {\sc Xuhua, L.} 1995.
\newblock The rationality and decidability of fuzzy implications.
\newblock In {\em Proc.\ of the 14th Int.\ Joint Conf.\ on Artificial
  Intelligence (IJCAI-95)}. MK, Montreal, Canada, 1910--1911.

\bibitem[\protect\citeauthoryear{Yager}{Yager}{1985}]{Yager85}
{\sc Yager, Ronald, R.} 1985.
\newblock Inference in multivalued logic system.
\newblock {\em Int. J. Man-Machine Studies\/}~{\em 23}, 27--44.

\bibitem[\protect\citeauthoryear{Zadeh}{Zadeh}{1965}]{Zadeh65}
{\sc Zadeh, L.~A.} 1965.
\newblock Fuzzy sets.
\newblock {\em Information and Control\/}~{\em 8,\/}~3, 338--353.

\end{thebibliography}

\end{document}